# A Benchmark Study by using various Machine Learning Models for Predicting Covid-19 trends

**D.Kamelesun, R.Saranya, P.Kathiravan**
Department of Computer Science, Central University of Tamil Nadu, India
saranya@cutn.ac.in

## Abstract

Machine learning and deep learning play vital roles in predicting diseases in the medical field. Machine learning algorithms are widely classified as supervised, unsupervised, and reinforcement learning. This paper contains a detailed description of our experimental research work in that we used a supervised machine-learning algorithm to build our model for outbreaks of the novel Coronavirus that has spreaded over the whole world and caused many deaths, which is one of the most disastrous Pandemics in the history of the world. The people suffered physically and economically to survive in this lockdown. This work aims to understand better how machine learning, ensemble, and deep learning models work and implements in the real dataset. In our work, we are going to analyze the current trend or pattern of the coronavirus and then predict the further future of the covid-19 confirmed cases or new cases by training the past Covid-19 dataset by using the machine learning algorithm such as Linear Regression, Polynomial Regression, K-nearest neighbor, Decision Tree, Support Vector Machine and Random forest algorithm are used to train the model.

The decision tree and the Random Forest algorithm perform better than SVR in this work. The performance of SVR and lasso regression are low in all prediction areas Because the SVR is challenging to separate the data using the hyperplane for this type of problem. So SVR mostly gives a lower performance in this problem. Ensemble (Voting, Bagging, and Stacking) and deep learning models(ANN) also predict well. After the prediction, we evaluated the model using MAE, MSE, RMSE, and MAPE. This work aims to find the trend/pattern of the covid-19.

# 1. Introduction

The First known Coronavirus was discovered in Wuhan, China. World Health Organisation(WHO) was informed about the case of an unknown virus caused in Wuhan in December 2019. The Symptoms of Coronavirus are cough and fever. The cough spreads this virus (Jignesh Chowdary et al., 2020; Liu, 2020; Su et al., 2022; Tiwari et al., 2020). This virus spread over most countries in the world. In our country, people are also terribly affected by this virus. In the last 20 years, As per the national center for Biotechnology Information record, they have recorded various coronaviruses such as SARS-Cov in 2002-2004, HINI Influenza in 2009, and MERS-Cov in Saudi Arabic in 2012, declared a Pandemic by WHO.

In this work, we took the covid-19 dataset from the Kaggle site. We built a Machine Learning model to predict the current trend/pattern of covid-19 and number of confirmed cases using machine-learning algorithms such as linear regression (Mojjada et al., 2021), Polynomial Regression (Shaikh et al., 2021; Yadav et al., 2020), SVR (Rivas-Perea et al., 2013), Lasso Regression (Jolliffe et al., 2003; Mojjada et al., 2021; Rustam et al., 2020), K-NN, Decision tree (DT), Random Forest, ensemble techniques such as voting, bagging, and stacking, and deep learning also used to build the model by current past dataset covid-19. After training, we evaluated the model using some evaluation metrics to find the accuracy of the models using MAE, MSE, RMSE, and MAPE (Kanimozhi et al., 2020; Liu, 2020; Verma & Pal, 2020). Following objectives were addressed in this paper.

**Objective 1**: To analyze the current trend of covid-19 confirmed cases by the past covid-19 data.
**Objective 2**: To predict the future of confirmed cases of covid-19 by training the model using the past covid-19 dataset.

The related works proposed methodologies, the model evaluations using various matrices, and the conclusion are in the rest of this paper.

# 2. Related work

| S.no | Authors | Methodology | Finding |
|---|---|---|---|
| 1 | (Gambhir et al., 2020) | SVR, Polynomial regression | Analyzes Current / patterns of covid-19 |
| 2 | (Mandayam et al., 2020) | Linear Regression and SVR | To predict the future number of positive cases |
| 3 | (Rustam et al., 2020) | Linear Regression LASSO Regression SVR | No. of newly infected cases, the no. of deaths, and the no. of recoveries in the next 10 days. |
| 4 | (Nikhil et al., 2021) | Polynomial Regression | Predicting the upcoming cases for next 25 days |
| 5 | (Liu, 2020) | Linear Regression Logistic Regression and RNN | predict pandemic data of the U.S |
| 6 | (Shaikh et al., 2021) | Linear Regression and Polynomial Regression | Analysis, prediction and Time series forecasting |
| 7 | (Painuli et al., 2021) | Random Forest and extra tree classifiers | one for predicting the chance of being infected and other for forecasting the number of positive cases |
| 8 | (Mojjada et al., 2021) | Linear Regression, Lasso Regression SVM, Exponential Smoothing, | The number of newly infected COVID 19 people, mortality rates and the recovered COVID-19 estimates in the next 10 days. |

| | | | |
|---|---|---|---|

# 3 Proposed methodology

In our Research, This research contains two main phases. The first phase is the training, and another one is testing. In the first phase, my dataset had many null or missing values, so we used pre-processing to clean the data. And then, we used Feature Scaling to modify the data from its original form because the machine can't be trained well with its original form of data (MinMax, AbslouteMax, Normalization, Standardization, Robust Scaling) (Boente & Salibian-Barrera, 2015; Lau & Baldwin, 2016; Lin et al., 2016; Storcheus, Dmitry; Rostamizadeh, Afshin; Kumar, 2015), so by the Feature scaling, we can modify the data for our convince to build the model. After the feature scaling, we segregated the data to train and test data. For training data, we prepared the model and the test data to evaluate the trained model, so we used the split function to split the data for training data is 80%, and test data is 20% of the datasets. For our experimental research work, we took the date and country attributes as the independent variable from the dataset. Both date and country attribute values are in String type. So we have to convert the value of the date attribute into numeric data by splitting the date into the date, month, and year and putting them into the different attributes, and then convert the country attributes data into numeric with the help of Label Encoding, which gives the numeric value for each country, Example India-1, USA-2, UK-3 like this it will create the numeric values for all the countries in the dataset. We analyzed the current trend or pattern of the coronavirus and then predict the further future of the covid-19 confirmed cases or new cases by Training the past Covid-19 dataset using the machine learning algorithm, ensembling models, and deep learning such as Linear Regression, Polynomial Regression, K-nearest neighbor, Lasso Regression, Decision Tree, Support Vector Machine, Decision tree, Random Forest algorithm, Artificial Neural Networks(ANN) and Improve the model evaluation by the Ensemble methods (Voting, Bagging, Stacking).

In the second phase, we first validated the model using the metrics of (MAE, MSE, RMSE, and MAPE) to observe the model's accuracy. The decision tree and the Random Forest algorithm perform better than these algorithms. The performance of SVR is low in all prediction areas because the SVR is challenging to separate the data using the hyperplane for this type of problem. So SVR mostly gives a low performance in this problem. What is the use of this paper? If the covid -19 comes, we have to know the current trend/pattern of the covid-19, so we need that current past dataset to train the model. After preparing the model to analyze the current trend of the covid-19, predict the future number of confirmed coronavirus cases in a day; how it works is shown in the below architecture Diagram (Fig. 1).

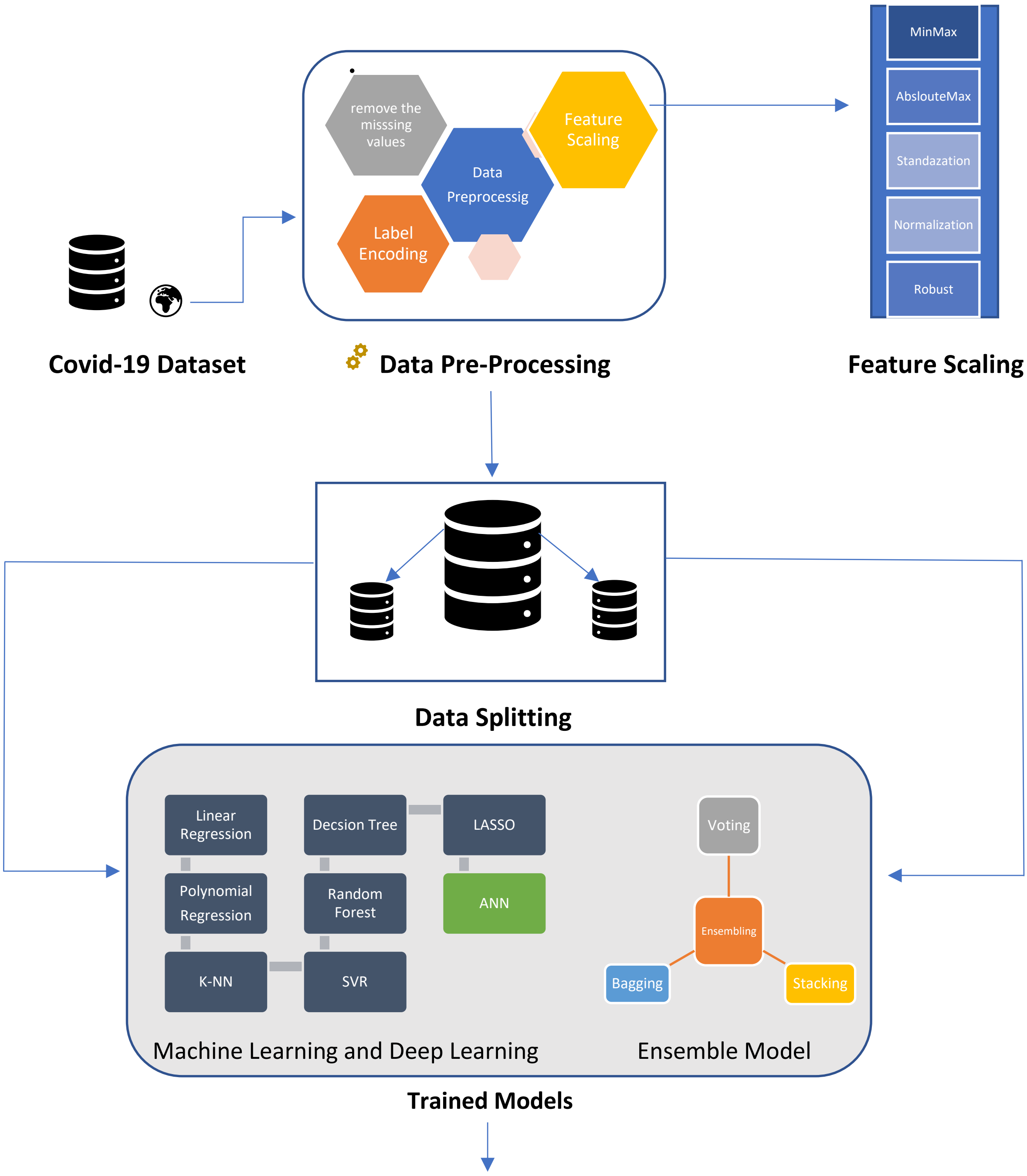

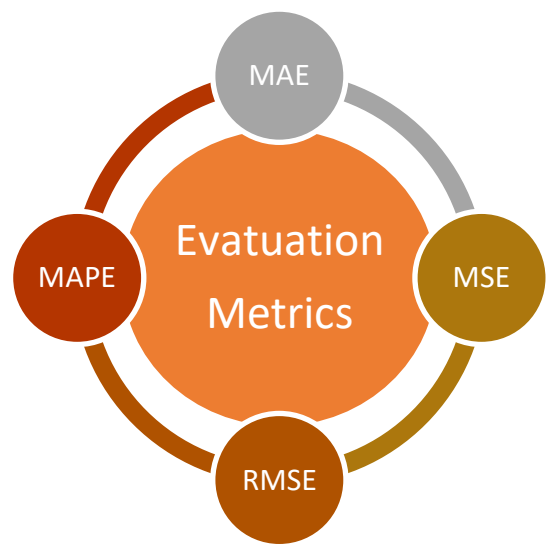

*Figure 1. Architecture Diagram*

# 3.1 Dataset Collection

## 1. Dataset Description

### Dataset-1

In this research, The dataset used was taken from the Kaggle Website (Gambhir et al., 2020), and contained the COVID-19 details. In this dataset, we have 201 countries' COVID -19 data. Each country has data from 15 February 2020 to 02nd 22 December 2021 in our dataset. And the attributes are **date**, **country**, how many people were affected by COVID-19 daily (**daily-new-cases**), the total number of affected people on that date(**cumulative-total-cases**), how many people are under treatment on that date(**active-cases**), and the total number of death cases (**cumulative-total- deaths**). How many people die daily? (**daily-new-deaths**), These are the attributes we have. The entire length of the dataset is 1,45,220.

| TOP 10 Countries | Cumulative total confirmed Cases | TOP 10 Countries | Cumulative total Death Cases |
|---|---|---|---|
| ***Global*** | ***264440620*** | ***Global*** | ***5249736*** |
| USA | 49716825 | USA | 806398 |
| INDIA | 34615757 | BRAZIL | 615225 |
| BRAZIL | 22118782 | INDIA | 470115 |
| UK | 10329063 | MEXICO | 294428 |
| RUSSIA | 9703107 | RUSSIA | 277640 |
| TURKEY | 8839891 | PERU | 201282 |
| FRANCE | 7773530 | UK | 145281 |
| IRAN | 6125596 | INDONESIA | 143850 |
| GERMANY | 6026796 | ITALY | 134003 |
| ARGENTINA | 5335310 | IRAN | 129988 |

## Dataset-2

In this research, The dataset used was taken from the Kaggle Website (Mandayam et al., 2020), and contained the COVID-19 details. In this dataset, we have 193 countries' COVID -19 data. Each country has data from 16 November 2020 to 12 September 2021 in our dataset. And the attributes are date, country, CountryAlphaCode, Confirmed cases maximum of 41million, Death cases maximum of 660k, Recoveries maximum of 31 million, ECR, GRTStringencyIndex, DaySinceFirstCases, DaySince100th Cases, ConfirmedPopPct, DeathPopPct, RecoveriesPopPct. These are the attributes we have. The total length of the dataset is 2952600.

| TOP 10 Countries | Cumulative total confirmed Cases | TOP 10 Countries | Cumulative total Death Cases | TOP 10 Countries | Cumulative total Recovery Cases |
|---|---|---|---|---|---|
| Global | **3.03868E+13** | Global | **6.18299E+11** | Global | **4.08442E+12** |
| US | 2.65162E+13 | US | 5.10454E+11 | Brazil | 2.67528E+12 |
| Brazil | 3.68155E+12 | Brazil | 1.01685E+11 | US | 1.34381E+12 |
| Canada | 68033025900 | China | 2712603648 | China | 45989925888 |
| China | 55937676288 | Canada | 1666076244 | India | 4859387857 |
| United Kingdom | 34461082300 | United Kingdom | 1067745075 | Russia | 1128064202 |
| India | 6529623206 | India | 87802534 | Turkey | 919259007 |
| Russia | 1562989024 | Mexico | 68604089 | Italy | 759237934 |
| France | 1511222838 | Peru | 55929784 | Colombia | 735029503 |
| Turkey | 1228359396 | Italy | 39566917 | Argentina | 711610324 |
| Spain | 1100245487 | France | 34621623 | Germany | 69524824 |

We had missing or null values in my dataset, so we eliminated the missing values from the dataset to build the good machine learning model.

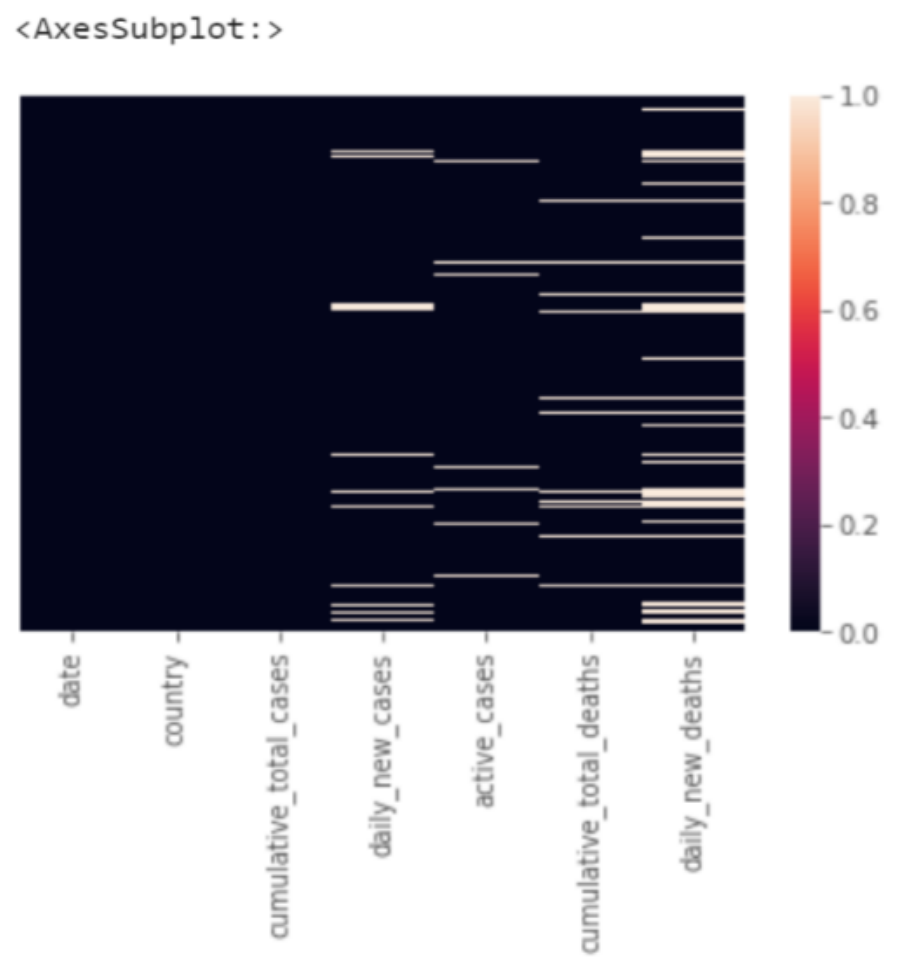

*Figure 2. Null values in dataset-1*

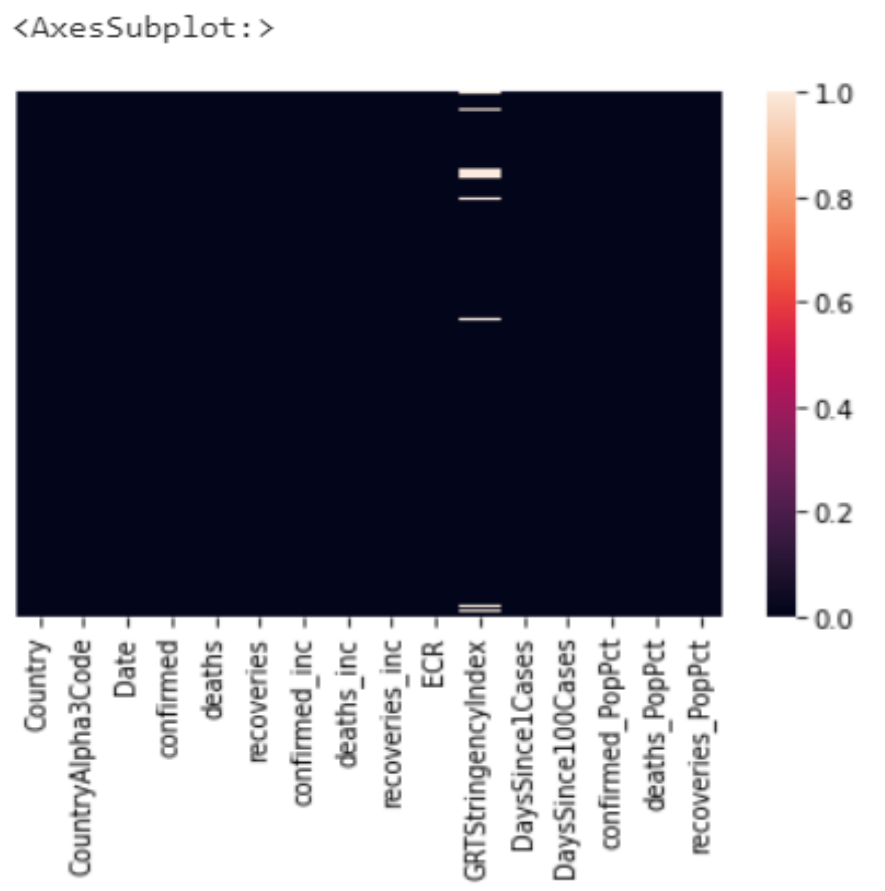

*Figure3. Null values of dataset-2*

I used the dropna() function to delete the missing value from the dataset. Figure 3 shows how many daily cases appear in each country. Below you will see how many null values are in the dataset by the graph.

| Dataset | Before Eliminating Null Values | After Eliminating Null Values |
|---|---|---|
| Dataset -1 | date 0<br>country 0<br>cumulative_total_cases 0<br>daily_new_cases 7491<br>active_cases 4599<br>cumulative_total_deaths 7227<br>daily_new_deaths 22791<br>dtype: int64 | date 0<br>country 0<br>cumulative_total_cases 0<br>daily_new_cases 0<br>active_cases 0<br>cumulative_total_deaths 0<br>daily_new_deaths 0<br>dtype: int64 |
| Dataset-2 | Country 0<br>CountryAlpha3Code 0<br>Date 0<br>confirmed 0<br>deaths 0<br>recoveries 0<br>confirmed_inc 0<br>deaths_inc 0<br>recoveries_inc 0<br>ECR 0<br>GRTStringencyIndex 130634<br>DaysSince1Cases 0<br>DaysSince100Cases 0<br>confirmed_PopPct 1200<br>deaths_PopPct 1200<br>recoveries_PopPct 1200<br>dtype: int64 | Country 0<br>CountryAlpha3Code 0<br>Date 0<br>confirmed 0<br>deaths 0<br>recoveries 0<br>confirmed_inc 0<br>deaths_inc 0<br>recoveries_inc 0<br>ECR 0<br>GRTStringencyIndex 0<br>DaysSince1Cases 0<br>DaysSince100Cases 0<br>confirmed_PopPct 0<br>deaths_PopPct 0<br>recoveries_PopPct 0<br>dtype: int64 |

*Table 2. Before and After eliminating null values*

# 2 Summary of the dataset

The covid-19 datasets are mixed of integer and real-valued attribute characteristics and this is used to evaluate the machine leaning models.

| Attributes | Dataset-1 | Dataset-2 |
|---|---|---|
| Data set characteristics | Covid-19 | Covid-19 |
| Attribute characteristics | Integer and alphabetic values | Integer and alphabetic values |
| No .of .Attributes | 07 | 16 |
| No .of .instance | 145220 | 2952600 |
| Missing values or errors present | yes | yes |

*Table3. Attributes and Details of the dataset*

# 3 Data Visualisation

## Dataset-1

The covid-19 global dataset created this Map, showing the Total confirmed cases of each country worldwide, separated by the colors depending upon the confirmed cases and the Total confirmed cases on each date. This plot was created by this (import plotly. express as px) (*Plotly Python Graphing Library*, n.d.; Preacher et al., 2006). This scale, from blue to yellow, refers to the range of covid-19 confirmed cases. If it shows dark blue, it has zero cases. If it shows yellow, it will range between 15 million to 20 million cases.

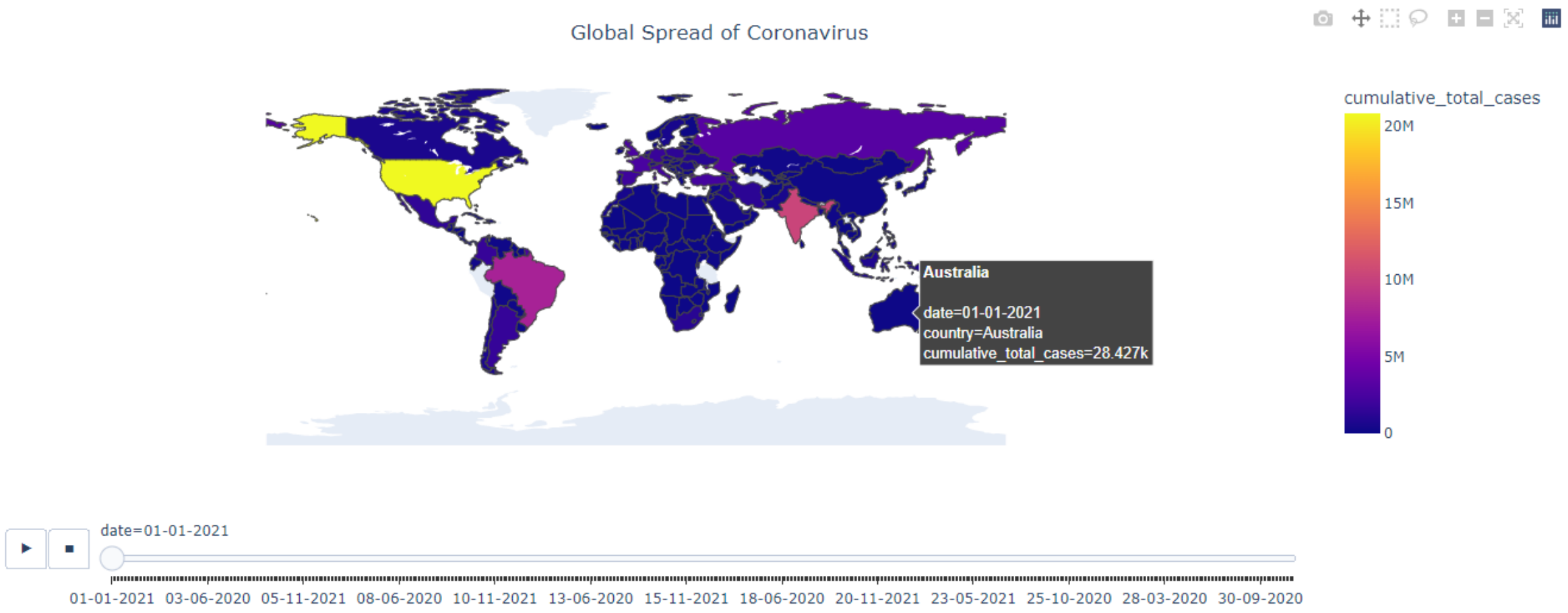

*Figure 4. Shows the Confirmed in all over the world dataset-1*

This Map also worked like the previous Confirmed cases Map. This Map shows the Total death cases of each country worldwide, each country separated by different colors depending upon the death cases and the Total death cases on each date.

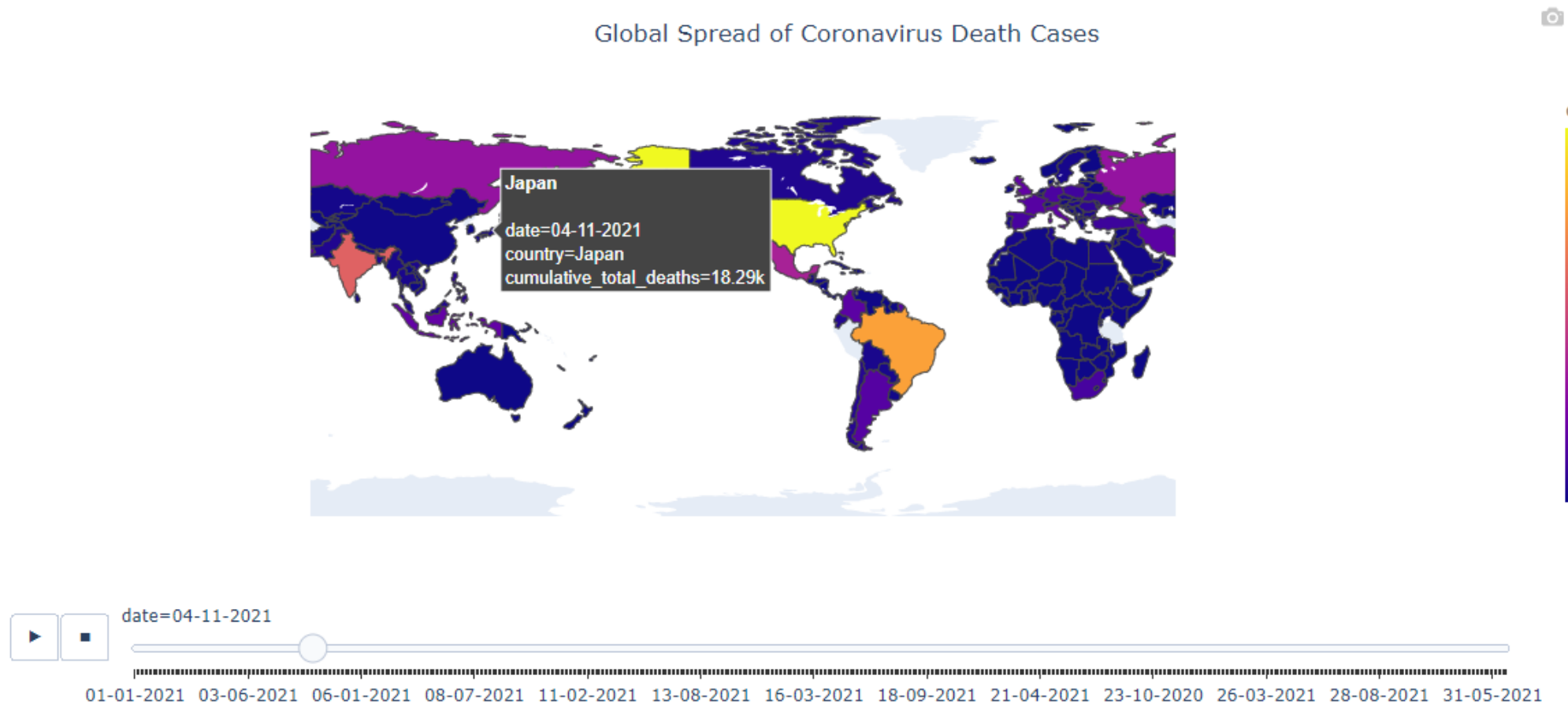

*Figure 5. Shows the Death Cases in all over World dataset-1*

## Dataset-2

The covid-19 global dataset created this Map, showing the Total confirmed cases of each country worldwide, separated by the colors depending upon the confirmed cases. This plot was created by this (import plotly. express as px). This scale, from blue to yellow, refers to the range of covid-19 confirmed cases. If it shows light sandals, it has zero cases. If it shows yellow, it will range between 1.5 billion to 2 billion cases.

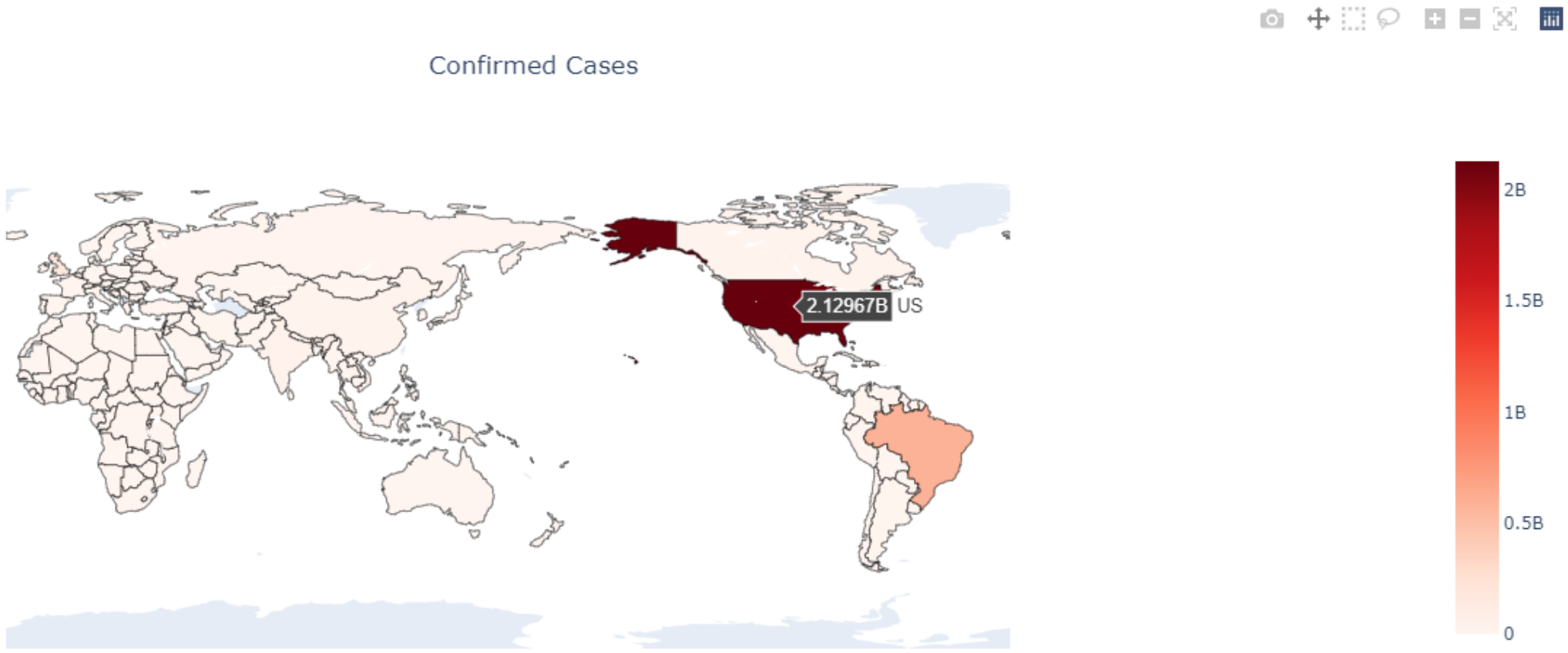

*Figure 6. Shows the Confirmed cases in all over world dataset-2*

---

This Map also worked like the previous Confirmed cases Map. This Map shows the Total death cases of each country worldwide, separated by the different colours depending upon the range of death cases in each country.

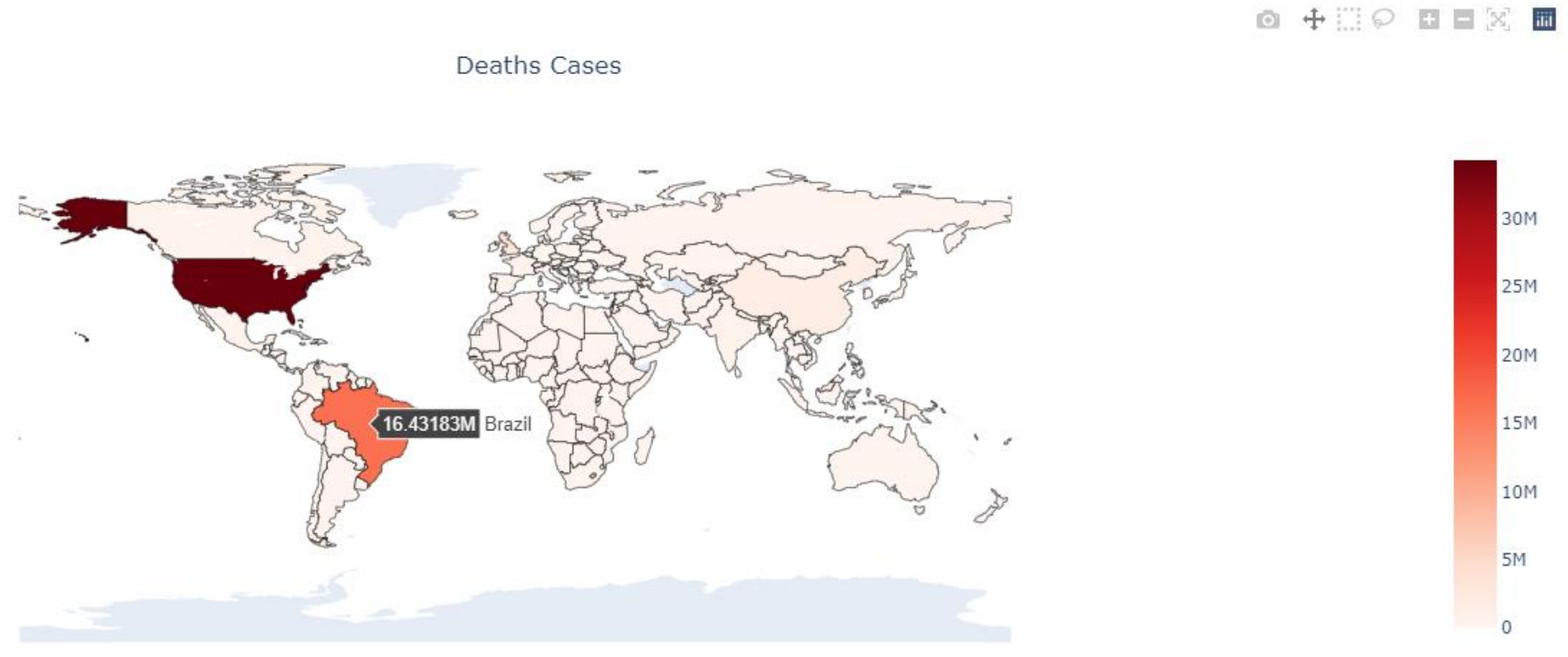

*Figure 7. Shows the Death cases in all over world dataset-2*

# Global Spread of covid-19 top 5 countries in the graph

## Dataset-1

We took five countries from this dataset to plot this graph with confirmed cases data on the y-axis and the date on the x-axis. From this graph, we can see the confirmed cases on the respective date for these countries.

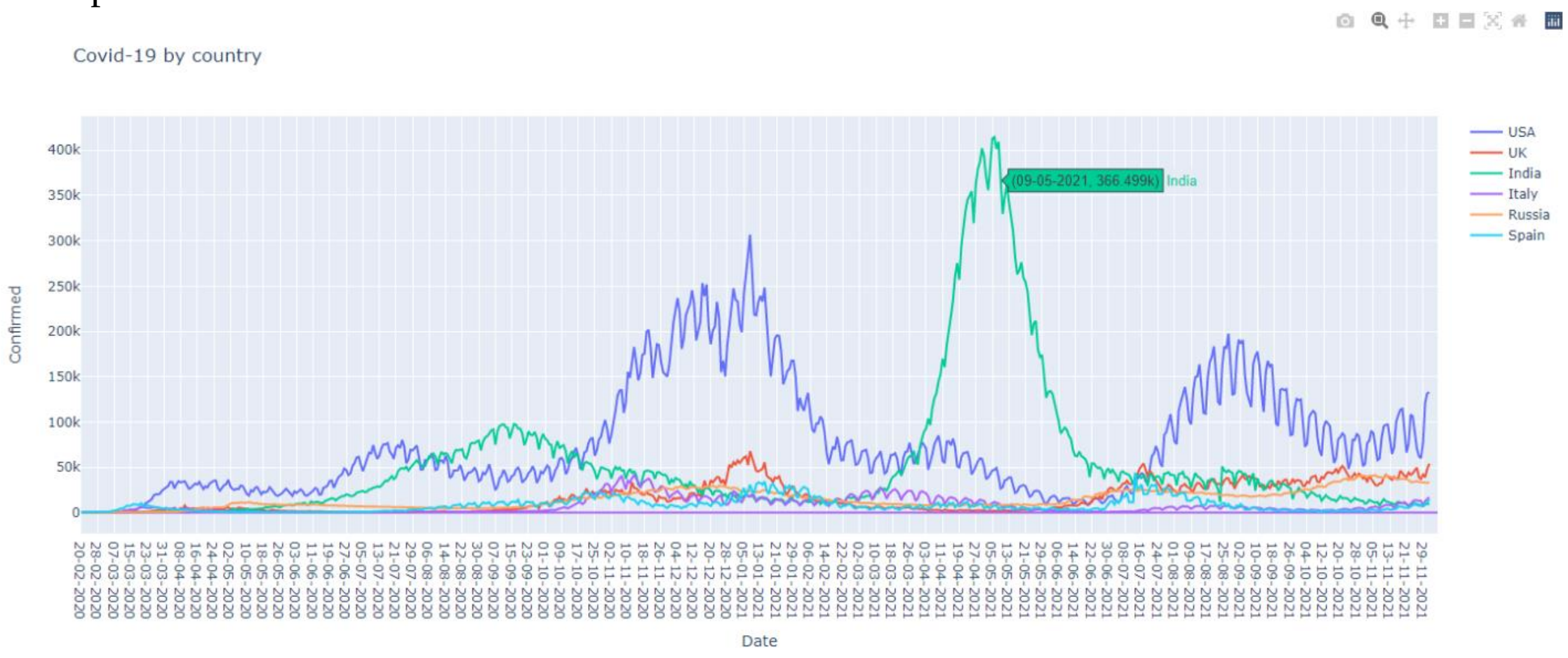

*Figure 8. Five countries confirmed cases data show in graph for dataset-1*

---

## Dataset-2

From this graph, you can see the data from the five countries of confirmed cases with the respective date(X-axis). Confirmed cases in Y-axis and date in X-axis. If we drag the mouse on the graph line, it will show the confirmed cases on respective dates for a country.

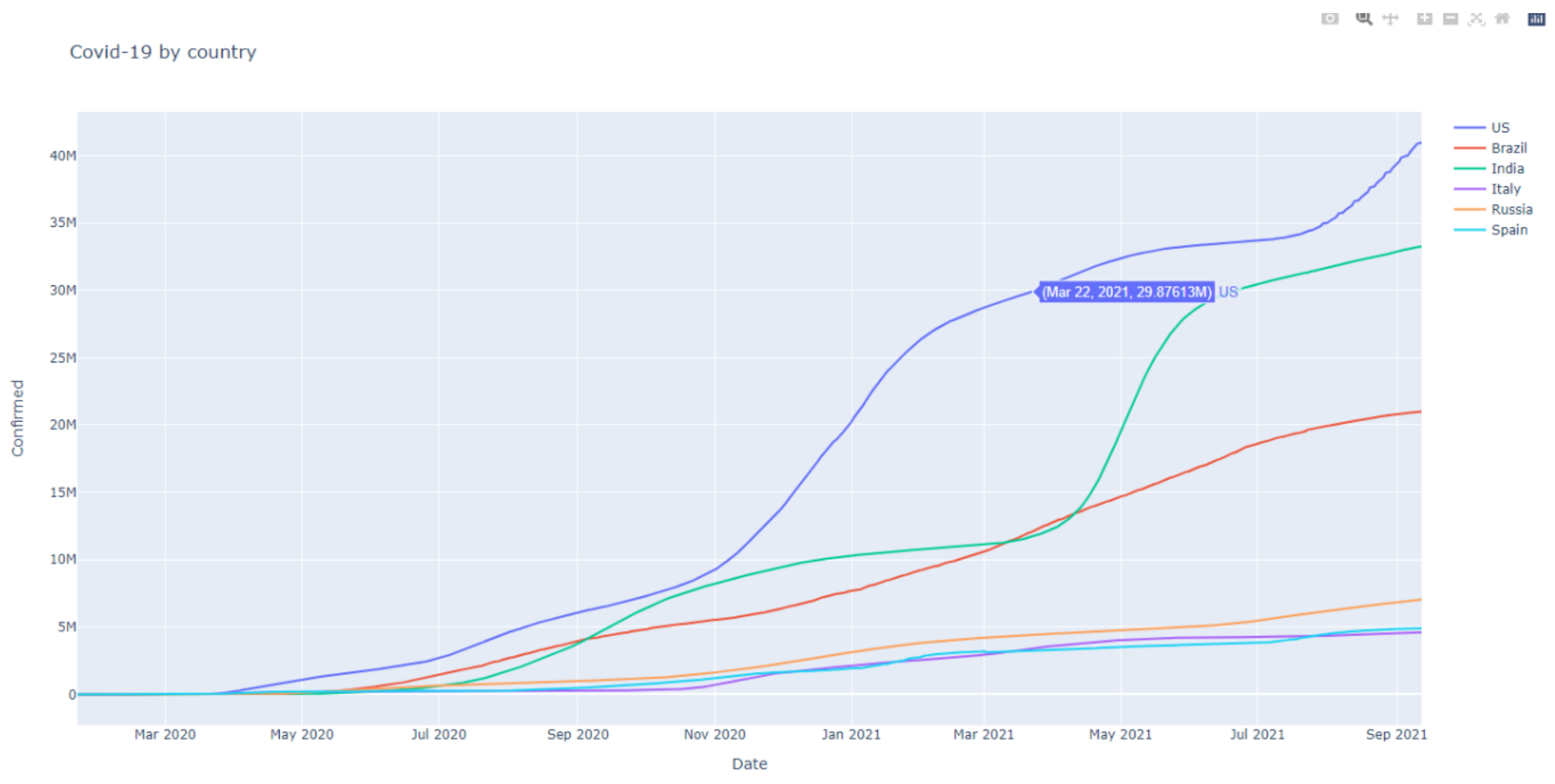

*Figure 9. Five countries confirmed cases data show in graph for dataset-1*

# 3.2.Data Pre-processing

## 1 Feature Scaling

### a) Absolute Maximum Scaling

Find the absolute maximum value in the column and divide all the values in the queue by that maximum value. Our data will lie between -1 and 1 (Mojjada et al., 2021; Rustam et al., 2020).

Absolute Maximum Scaling= Each value in the column / Maximum value of the column (1)

In our experimental research work, we used this Feature scaling for the dependent and independent variables to enhance the model by changing the values of the dataset into the range of -1 to 1 because these data will understand by engines easily. If I do the model training with my original data, the model may not be trained well.

### b) MinMax Scaling

Subtract each column value by the minimum value of the column and then divide this by the subtracted value of the maximum and minimum value of the column. By this method, our data will lie between 0 and 1. if we have negative values in the dataset. This Scaling will compress all the inliers in the narrow range [0, 0.005] .

MinMax Scaling= (X - Min(X)) / (Max(X) - Min(X)) (2)

We used this feature scaling to enhance the model of my experimental research work by changing the values of the dependent and independent variables into the range -1 to 1 for machine-readable (*Sklearn.Preprocessing.Minmax_scale — Scikit-Learn 1.1.3 Documentation*, n.d.) .

### c) Normalization

In this case, we used the average value of the column instead of the minimum value.

Normalization Scaling = (X - Mean(X)) / (Max(X) - Min(X)) (3)

In my Experimental research work, I have the columns with different values, not in sequential order. I used this Scaling to enhance the model by changing the values of the dataset into the range of 0 to 1 (Lin et al., 2016; Storcheus, Dmitry; Rostamizadeh, Afshin; Kumar, 2015).

### d) Standardization

Each column value subtracts the average value of the column and then is divided by the Standard deviation value and is not restricted to a specific range. In this experimental research work. We used this to enhance the model by changing the values of the dataset into the range of [0, 1]. (*Sklearn.Preprocessing.StandardScaler — Scikit-Learn 1.1.3 Documentation*, n.d.).

Standardization Scaling = (X – Mean(X)) / σ (4)

### e) Robust Scaling

Subtract the data points with the median value and divide them by the Inter Quartile Range value (IQR). IQR means the difference between the dataset's first and third quartile (Barros & Hirakata, 2003; Boente & Salibian-Barrera, 2015).

Robust Scaling = X – Median(X) / IQR. (5)

## 2 Date Splitting

In our dataset, The date in the String value can't be readable by the machine, so we have changed the date into an understandable format by splitting the date into three columns date, day, and year.

| | date | country | cumulative_total_cases | daily_new_cases | active_cases | cumulative_total_deaths | daily_new_deaths | Dates | Month | Year | Country |
|---|---|---|---|---|---|---|---|---|---|---|---|
| **10** | 25-02-2020 | Afghanistan | 1 | 0.0 | 1.0 | 0.0 | 0.0 | 25 | 2 | 2020 | 0 |
| **11** | 26-02-2020 | Afghanistan | 1 | 0.0 | 1.0 | 0.0 | 0.0 | 26 | 2 | 2020 | 0 |
| **12** | 27-02-2020 | Afghanistan | 1 | 0.0 | 1.0 | 0.0 | 0.0 | 27 | 2 | 2020 | 0 |
| **13** | 28-02-2020 | Afghanistan | 1 | 0.0 | 1.0 | 0.0 | 0.0 | 28 | 2 | 2020 | 0 |
| **14** | 29-02-2020 | Afghanistan | 1 | 0.0 | 1.0 | 0.0 | 0.0 | 29 | 2 | 2020 | 0 |
| **...** | ... | ... | ... | ... | ... | ... | ... | ... | ... | ... | ... |
| **145216** | 28-11-2021 | Zimbabwe | 133991 | 40.0 | 631.0 | 4705.0 | 0.0 | 28 | 11 | 2021 | 200 |
| **145217** | 29-11-2021 | Zimbabwe | 134226 | 235.0 | 817.0 | 4706.0 | 1.0 | 29 | 11 | 2021 | 200 |
| **145218** | 30-11-2021 | Zimbabwe | 134625 | 399.0 | 1171.0 | 4707.0 | 1.0 | 30 | 11 | 2021 | 200 |
| **145219** | 01-12-2021 | Zimbabwe | 135337 | 712.0 | 1846.0 | 4707.0 | 0.0 | 12 | 1 | 2021 | 200 |
| **145220** | 02-12-2021 | Zimbabwe | 136379 | 1042.0 | 2843.0 | 4707.0 | 0.0 | 12 | 2 | 2021 | 200 |

119189 rows × 11 columns

## 3 Label Encoding

Converting the alphabetic values into a numeric form converts them into a machine-readable now machine-learning algorithm that can easily understand it(*Sklearn.Preprocessing.LabelEncoder — Scikit-Learn 1.1.3 Documentation*, n.d.) . Because it transformed into categorical values, for example, India, USA, U.K… by this method, it will change like 0, 1, 2, … and so on. In our Experimental research work, we used the country column. Because the machine learning algorithm can't read the alphabetic values, we used this method to change them into categorical values.

| | date | country | cumulative_total_cases | daily_new_cases | active_cases | cumulative_total_deaths | daily_new_deaths | Datess | Month | Year | Country |
|---|---|---|---|---|---|---|---|---|---|---|---|
| 10 | 25-02-2020 | Afghanistan | 1 | 0.0 | 1.0 | 0.0 | 0.0 | 25 | 2 | 2020 | 0 |
| 11 | 26-02-2020 | Afghanistan | 1 | 0.0 | 1.0 | 0.0 | 0.0 | 26 | 2 | 2020 | 0 |
| 12 | 27-02-2020 | Afghanistan | 1 | 0.0 | 1.0 | 0.0 | 0.0 | 27 | 2 | 2020 | 0 |
| 13 | 28-02-2020 | Afghanistan | 1 | 0.0 | 1.0 | 0.0 | 0.0 | 28 | 2 | 2020 | 0 |
| 14 | 29-02-2020 | Afghanistan | 1 | 0.0 | 1.0 | 0.0 | 0.0 | 29 | 2 | 2020 | 0 |
| ... | ... | ... | ... | ... | ... | ... | ... | ... | ... | ... | ... |
| 145216 | 28-11-2021 | Zimbabwe | 133991 | 40.0 | 631.0 | 4705.0 | 0.0 | 28 | 11 | 2021 | 200 |
| 145217 | 29-11-2021 | Zimbabwe | 134226 | 235.0 | 817.0 | 4706.0 | 1.0 | 29 | 11 | 2021 | 200 |
| 145218 | 30-11-2021 | Zimbabwe | 134625 | 399.0 | 1171.0 | 4707.0 | 1.0 | 30 | 11 | 2021 | 200 |
| 145219 | 01-12-2021 | Zimbabwe | 135337 | 712.0 | 1846.0 | 4707.0 | 0.0 | 12 | 1 | 2021 | 200 |
| 145220 | 02-12-2021 | Zimbabwe | 136379 | 1042.0 | 2843.0 | 4707.0 | 0.0 | 12 | 2 | 2021 | 200 |

119189 rows × 11 columns

# 4 Model Building

## 4.1 Machine learning

### a) Linear Regression

This algorithm is used to find the changes of the dependent variable according to the independent variable and shows the difference between dependent and independent variables, known as Linear regression (Mojjada et al., 2021; Rustam et al., 2020; Sardinha & Catalán, 2018; Yadav et al., 2020). It plots the straight line and covers most of the data points in the dependent variable. For continuous/real/numeric variables like sales, salary, age, and product price, among others, linear regression makes predictions.

Two types of linear regression

Simple linear regression - predict the value by a single independent variable.

Multiple linear regression - predict the value by more than one independent variable.

$$\mathbf{Y = mX + b} \quad (6)$$

Y is the dependent variable

X is the independent variable

m is the slope of the line

b is the intercept

In this experimental research work, we used Multiple Linear Regression (Rath et al., 2020; Sardinha & Catalán, 2018). We took the date and country as the independent variable and the confirmed cases as the dependent variable. By this algorithm, we can predict the confirmed cases of covid-19 by giving the value of the input of the independent variable to the trained model and predicting the output, respectively.

## b) Polynomial Regression

It is also known as the Multiple Linear Regression Special Case in Machine Learning. Because transforming the equation for multiple linear regression into polynomial regression by adding specific polynomial terms. It is a linear model that has been modified in several ways to improve accuracy. The training dataset for polynomial regression is non-linear. To fit the complex and non-linear functions and datasets instead of using linear regression. This statement leads to the polynomial regression: transformed original features into polynomial features of the essential degree (2,3,..,n) and then used in a linear model (Nikhil et al., 2021; Shaikh et al., 2021; Yadav et al., 2020).

$$y = b_0 + b_1x + b_2x_2 + b_3x_3 + .... + b_nx_n \quad (7)$$

This formula also works like the linear regression formula but changes the $n^{th}$ degree to plot the line closure to the data points.

This algorithm also works like linear regression, but we can change the predicted output by changing the degree value of polynomial regression. By this change, we can give the optimized predicted output for the given input.

## c) K-NN

**KNN -** K-Nearest Neighbour

The K-NN algorithm first finds the similarity between the new case and the existing cases, and then it places the new case in a category that is quite similar to the existing categories. After storing all of the previous data, a new data point is categorized using the K-NN Algorithm based on the similarity or distance of the data points. This model will quickly recognize and place new cases in the most related category.

The Euclidean distance between the data points will be calculated. The distance between two points is known as the Euclidean distance. The data points are classified based on this Euclidean distance.

$$d=\sqrt{(x2-x1)^2+(y2-y1)^2} \quad (8)$$

By this example, you can understand the formula,

$$d=\sqrt{(age2-age1)^2+(gender2-gender1)^2} \quad (9)$$

This algorithm is typically used for classification problems, but we used it in a regression-type problem this time, and it predicted the confirmed cases but did not perform well.

## d) Decision tree

A decision tree is one of the most popular supervised learning methods to solve classification and regression problems (Deng et al., 2019; Safavian & Landgrebe, 1991). Structured classifier,
where internal nodes stand for a dataset's features, branches for the decision-making process, and each leaf node for the classification result (Alghamdi & Alfalqi, 2015; Verma & Pal, 2020). It is used to split the datasets into tree-like structures for decision-making.

The Decision Node and Leaf Node are the two decision tree nodes. While Leaf nodes are the results of decisions and do not have any more branches, Decision nodes are used to create decisions and have numerous branches. The CART algorithm divides a node into sub-nodes based on the Gini Index value. It starts with the training set as a root node, and after successfully splitting it in two, it breaks the subsets using the same logic and then splits the sub-subsets again, recursively, until it discovers that further splitting will not result in any pure sub-nodes.

We used this algorithm in my experimental research work to build the model and predict the confirmed cases of the covid-19. The evaluation metrics proved that this algorithm worked well for our experimental dataset.

## e) Random Forest

Random Forest is a classifier. It also works like the decision tree algorithm, but it uses many decision trees on different subsets of the input dataset. Each tree has the predicted output by voting averages of the results to increase the predicted accuracy of the model (Ankit & Saleena, 2018; Hossain et al., 2021; Painuli et al., 2021; Pokharel & Deardon, 2014). This algorithm can improve the accuracy and performance of the model from the decision tree model. More trees in the random forest result in increased accuracy and high performance and prevent (or) avoid the problem of overfitting. In my experimental research work, I used this algorithm to build the model and predict the confirmed cases of the covid-19. This algorithm also worked well than the decision tree because and proved by the evaluation metrics.

## f) SVM

Support Vector Machine (SVM) is a popular Supervised Learning algorithm used for classification and regression problems. However, it is primarily used in Machine Learning for Classification problems. The SVM algorithm aims to find the best line or decision boundary for categorizing n-dimensional space to place new data points in the correct category easily. A hyperplane is the best decision boundary (Education, 2021; Oumina et al., 2020; Rustam et al., 2020; Suhasini et al., n.d.).

Support Vector Regression (SVR) is a supervised learning algorithm for predicting discrete values (Rivas-Perea et al., 2013). Support Vector Regression operates on the same principles as SVMs. SVR's basic concept is to find the best fit line. The best fit line in SVR is the hyperplane with the most significant number of points. This algorithm was used to create the model and predict the confirmed cases of the covid-19. However, the SVR performance is also low because it is challenging to construct the hyperplane for all data points in the dataset.

## 4.2 Ensembling Models

### a) Voting

Voting classifiers and regressors are both ensemble methods; the predictions of these models are simply an aggregation of ensemble predictions (Ankit & Saleena, 2018; Hammar et al., 2019). An ensemble is a collection of predictors. As a result, these models are made up of multiple predictors. The model aggregates each of these predictors' predictions into a final one (Agnihotri et al., 2019). In this experimental research work, we used voting for the regression model. I took the three machine learning models, the K-NN model, the Decision tree model, and the Random Forest Model; we calculated the result by averaging those outputs from these model outputs.

### b) Stacking

Stacking is a popular ensemble modeling technique for predicting multiple nodes to generate a new model and enhance model performance (Kim et al., 2021; Verma & Pal, 2020). We used the stacking technique in my experimental research to improve the model. We combined the three machine learning models, the K-NN model, the Decision tree model, and Random Forest Model by combining these models to build the new model and improve the model performance.

### c) Bagging

Bagging, also known as Bootstrap aggregation, is an ensemble learning technique that helps machine learning algorithms to improve their performance and accuracy. It is used to deal with bias-variance trade-offs and reduces a prediction model's variance. Bagging prevents data overfitting and is used in both regression and classification models, particularly for decision tree algorithms (Dong & Qian, 2022).

In this experimental research work, we are going to use bagging for the regression model to enhance the previously used model performance. We took the Decision tree algorithm to improve the model performance of this algorithm.

## 4.3 Deep Learning

### a) ANN

Artificial Neural Networks (ANN) are multi-layer fully-connected neural nets. They are made up of an input layer, several hidden layers, and an output layer. Every node in one layer is linked to every node. A structure like the human brain Artificial neural networks, like the human brain, have neurons that are interconnected to one another in various layers of the networks. These neurons are called nodes (Kathiravan & Saranya, 2021; Nan & Gao, 2018; Oumina et al., 2020; Ozturk et al., 2020).

In this Experimental research work, we used this Deep learning Algorithm to build the model and predict the confirmed cases of covid-19. The Performance of this Algorithm is also very well for our experimental dataset.

# 5.Evaluation

| USED ALGORITHMS | SCALING | **Dataset—1** | | | | **Dataset--2** | | | |
|---|---|---|---|---|---|---|---|---|---|
| | | MAE | MSE | RMSE | MAPE | MAE | MSE | RMSE | MAPE |
| Linear Regression | Absolute Maximum | 0.00773 | 0.00059 | 0.02430 | 0.77380 | 0.00327 | 0.00019 | 0.01381 | 0.32732 |
| | Min Max | 0.00786 | 0.00076 | 0.02759 | 0.78608 | 0.00318 | 0.00018 | 0.01357 | 0.31863 |
| | Normalization | 0.27527 | 0.13680 | 0.36987 | 27.5272 | 0.44087 | 0.22091 | 0.47001 | 44.0871 |
| | Standardization | 0.29441 | 0.95309 | 0.97626 | 29.4419 | 0.24278 | 1.02367 | 1.01176 | 24.2787 |
| | Robust | 3.97267 | 173.280 | 13.1636 | 397.267 | 7.86555 | 1051.98 | 32.4343 | 786.555 |
| Polynomial Regression | Absolute Maximum | 0.00772 | 0.00058 | 0.02426 | 0.77240 | 0.00322 | 0.00018 | 0.01374 | 0.32228 |
| | Min Max | 0.00782 | 0.00075 | 0.02755 | 0.78286 | 0.00313 | 0.00018 | 0.01351 | 0.31378 |
| | Normalization | 0.27342 | 0.13595 | 0.36872 | 27.3423 | 0.43739 | 0.21898 | 0.46796 | 43.7397 |
| | Standardization | 0.29424 | 0.94983 | 0.97459 | 29.4241 | 0.23882 | 1.01550 | 1.00772 | 23.8825 |
| | Robust | 3.96498 | 172.826 | 13.1463 | 396.498 | 7.73825 | 1041.97 | 32.2795 | 773.825 |
| K-NN | Absolute Maximum | 0.00779 | 0.00076 | 0.02758 | 0.77987 | 0.00339 | 0.00020 | 0.01441 | 0.33969 |
| | Min Max | 0.00735 | 0.00065 | 0.02563 | 0.73524 | 0.00308 | 0.00018 | 0.01342 | 0.30884 |
| | Normalization | 0.20626 | 0.10418 | 0.32277 | 20.6264 | 0.29956 | 0.14464 | 0.38032 | 29.9567 |
| | Standardization | 0.28259 | 0.95238 | 0.97590 | 28.2595 | 0.23513 | 1.00703 | 1.00351 | 23.5139 |
| | Robust | 3.83587 | 184.926 | 13.5987 | 383.587 | 7.65189 | 1051.12 | 32.4210 | 765.189 |
| SVR | Absolute Maximum | 0.09660 | 0.00961 | 0.09803 | 9.66083 | 0.09847 | 0.00979 | 0.09894 | 9.84745 |
| | Min Max | 0.09678 | 0.00979 | 0.09895 | 9.6785 | 0.09860 | 0.00979 | 0.09897 | 9.86086 |
| | Normalization | 0.23199 | 0.14200 | 0.37683 | 23.1999 | 0.37507 | 0.28580 | 0.53461 | 37.5072 |
| | Standardization | 0.23072 | 1.03625 | 1.01796 | 23.072 | 0.20443 | 0.85370 | 0.92396 | 20.4438 |
| | Robust | 2.72609 | 199.169 | 14.1127 | 272.60 | 4.75392 | 948.336 | 30.7950 | 475.392 |
| Decision Tree | Absolute Maximum | 0.00122 | 6.42435 | 0.00801 | 0.12245 | 0.00050 | 2.37978 | 0.00487 | 0.05063 |
| | Min Max | 0.00127 | 7.28606 | 0.00853 | 0.12702 | 0.00049 | 2.23365 | 0.00472 | 0.04979 |
| | Normalization | 0.12954 | 0.12982 | 0.36030 | 12.954 | 0.12124 | 0.12215 | 0.34950 | 12.1241 |
| | Standardization | 0.04880 | 0.11049 | 0.33241 | 4.88080 | 0.03811 | 0.17476 | 0.41805 | 3.81192 |
| | Robust | 0.65679 | 16.8198 | 4.10119 | 65.6796 | 1.26035 | 192.319 | 13.8679 | 126.035 |
| Random Forest | Absolute Maximum | 0.00113 | 3.71692 | 0.00609 | 0.11386 | 0.00046 | 1.89923 | 0.00435 | 0.04655 |
| | Min Max | 0.00116 | 4.25899 | 0.00652 | 0.11669 | 0.00046 | 2.23436 | 0.00472 | 0.04675 |
| | Normalization | 0.13557 | 0.08211 | 0.28656 | 13.5574 | 0.12545 | 0.07804 | 0.27936 | 12.5452 |
| | Standardization | 0.04479 | 0.07184 | 0.26804 | 4.47991 | 0.03484 | 0.12288 | 0.35054 | 3.48449 |
| | Robust | 0.62770 | 16.1979 | 4.02466 | 62.7707 | 1.13672 | 139.042 | 11.7916 | 113.672 |
| Voting Ensemble | Absolute Maximum | 0.00288 | 9.84888 | 0.00992 | 0.28810 | 0.00127 | 5.05432 | 0.00710 | 0.12745 |
| | Min Max | 0.00278 | 9.67740 | 0.00983 | 0.27831 | 0.00116 | 3.73179 | 0.00610 | 0.11633 |
| | Normalization | 0.15719 | 0.08449 | 0.29067 | 15.7198 | 0.18187 | 0.08527 | 0.29202 | 18.1877 |
| | Standardization | 0.10422 | 0.15918 | 0.39897 | 10.4228 | 0.08832 | 0.17096 | 0.41348 | 8.83253 |
| | Robust | 1.45676 | 35.3470 | 5.94533 | 145.676 | 2.82742 | 176.511 | 13.2857 | 282.742 |
| Bagging Ensemble | Absolute Maximum | 0.00115 | 3.58553 | 0.00598 | 0.11504 | 0.00047 | 2.39693 | 0.00489 | 0.04786 |
| | Min Max | 0.00130 | 7.81970 | 0.00884 | 0.13041 | 0.00047 | 2.17509 | 0.00466 | 0.04714 |
| | Normalization | 0.13660 | 0.07791 | 0.27912 | 13.6604 | 0.12664 | 0.07389 | 0.27183 | 12.6646 |
| | Standardization | 0.04498 | 0.07324 | 0.27064 | 4.49816 | 0.03515 | 0.11348 | 0.33686 | 3.51532 |
| | Robust | 0.63517 | 16.8967 | 4.11056 | 63.5177 | 1.07096 | 84.6469 | 9.20037 | 107.096 |
| Stacking Ensemble | Absolute Maximum | 0.00072 | 1.78910 | 0.00422 | 0.07217 | 0.00035 | 6.49338 | 0.00254 | 0.03549 |
| | Min Max | 0.00067 | 1.53106 | 0.00391 | 0.06747 | 0.00023 | 3.76875 | 0.00194 | 0.02359 |
| | Normalization | 0.09456 | 0.02263 | 0.15046 | 9.4561 | 0.07794 | 0.01525 | 0.12349 | 7.79464 |
| | Standardization | 0.01444 | 0.00667 | 0.08172 | 1.44478 | 0.01977 | 0.01457 | 0.12073 | 1.97717 |
| | Robust | 0.23091 | 1.68392 | 1.29766 | 23.0917 | 0.60178 | 19.0907 | 4.36929 | 60.1785 |
| ANN | Absolute Maximum | 0.00524 | 0.00067 | 0.02597 | 0.52460 | 0.00222 | 0.00022 | 0.01500 | 0.22281 |
| | Min Max | 0.00521 | 0.00067 | 0.02597 | 0.52182 | 0.00203 | 0.00017 | 0.01328 | 0.20345 |
| | Normalization | 0.16492 | 0.16498 | 0.40618 | 16.4920 | 0.34391 | 0.34419 | 0.58668 | 34.3914 |
| | Standardization | 0.19644 | 1.09775 | 1.04773 | 19.644 | 0.15029 | 0.87252 | 0.93408 | 15.0297 |
| | Robust | 2.76430 | 214.265 | 14.6378 | 276.43 | 4.74052 | 948.755 | 30.8018 | 474.052 |

# 6. Conclusion

In this experimental research work, we have done predictive analysis by proposing a new model to predict the recent covid-19 cases. Before training, we made some changes in the dataset for our convenience by pre-processing techniques such as Dropna() function to delete the null values in the dataset to avoid wrong prediction performance and separate the date column into the day, month and year because the string can not load in the machine learning. And then we used scaling to change the data in the range between 0 to 1 for easy understanding of the Machine Learning model because the original data is tough to train and predict, so we made these changes in the original data from 0 to 1, such as Absolute maximum, Min-Max, Normalization, Standardization, and Robust. Then, in the Data visualization phase, we used maps and graphs for an easy understanding of the data those we used in this experimental research. After that, we built the model using different machine learning algorithms such as Linear Regression (L.R.), Polynomial Regression (PR), K-nearest neighbor (KNN), Decision Tree (D.T.), Random Forest (R.F.), Lasso Regression (LR), Support Vector Regression (SVR), and we used the ensembling model also to improve the model performance. In ensembling techniques, we used voting, Stacking, and Bagging.

In deep learning Algorithms, the evaluation metrics such as MAE, MSE, RSME, and MAPE are used in our prediction model. Out of this decision tree and Random Forest algorithm performed better than SVR and all other algorithms. This model helps in predicting trends and patterns of covid-19.